\pdfoutput=1

\documentclass[11pt]{article}

\usepackage[]{acl}

\usepackage{times}
\usepackage{latexsym}
\usepackage[T1]{fontenc}
\usepackage[utf8]{inputenc}
\usepackage{microtype}
\usepackage{inconsolata}
\usepackage{graphicx}
\usepackage{booktabs, arydshln}
\usepackage{caption}
\usepackage{subcaption}
\usepackage{graphicx}
\usepackage{amssymb, bm}
\usepackage{tcolorbox}
\usepackage{comment}
\usepackage{color, colortbl}
\usepackage{siunitx}
\usepackage{multirow}
\usepackage{cleveref}
\usepackage{pifont}

\makeatletter
\def\adl@drawiv#1#2#3{%
        \hskip.5\tabcolsep
        \xleaders#3{#2.5\@tempdimb #1{1}#2.5\@tempdimb}%
                #2\z@ plus1fil minus1fil\relax
        \hskip.5\tabcolsep}
\newcommand{\cdashlinelr}[1]{%
  \noalign{\vskip\aboverulesep
           \global\let\@dashdrawstore\adl@draw
           \global\let\adl@draw\adl@drawiv}
  \cdashline{#1}
  \noalign{\global\let\adl@draw\@dashdrawstore
           \vskip\belowrulesep}}
\makeatother

\title{AAdaM at SemEval-2024 Task 1: Augmentation and Adaptation \\for Multilingual Semantic Textual Relatedness}

\author{Miaoran Zhang\textsuperscript{1} \,
 Mingyang Wang\textsuperscript{2,3} \, Jesujoba O. Alabi\textsuperscript{1} \vspace{3px} \, \textbf{Dietrich Klakow}\textsuperscript{1} \\
\textsuperscript{1}Saarland University, Saarland Informatic Campus \\
\textsuperscript{2}Bosch Center for AI,
\textsuperscript{3}LMU Munich\\
 {\tt  mzhang@lsv.uni-saarland.de}
}

\begin{document}
\maketitle
\begin{abstract}
This paper presents our system developed for the SemEval-2024 Task 1: Semantic Textual Relatedness for African and Asian Languages. The shared task aims at measuring the semantic textual relatedness between pairs of sentences, with a focus on a range of under-represented languages. In this work, we propose using machine translation for data augmentation to address the low-resource challenge of limited training data. Moreover, we apply task-adaptive pre-training on unlabeled task data to bridge the gap between pre-training and task adaptation. For model training, we investigate both full fine-tuning and adapter-based tuning, and adopt the adapter framework for effective zero-shot cross-lingual transfer. We achieve competitive results in the shared task: our system performs the best among all ranked teams in both subtask A (supervised learning) and subtask C (cross-lingual transfer).\footnote{Our code: \url{https://github.com/uds-lsv/AAdaM}}
\end{abstract}

\section{Introduction}
Semantic Textual Relatedness (STR) measures the closeness of meaning between two linguistic units, such as a pair of words or sentences~\citep{budanitsky1999lexical, mohammad2012distributional}. For example, one can easily tell that \textit{``I like playing games''} is more semantically related to \textit{``The game is fun''} rather than \textit{``The weather is good''}, which largely depends on their lexical semantic relation and topic consistency. Semantic Textual Similarity (STS), a closely related concept, indicates whether two units have a paraphrasing relation. The difference between these two concepts is clarified in \citet{abdalla-etal-2023-makes}: while similar pairs are also related, the reverse is not necessarily true.

In stark contrast to the extensive research on STS~\citep{gao-etal-2021-simcse, chuang-etal-2022-diffcse, zhang-etal-2022-mcse, seonwoo-etal-2023-ranking}, exploration of STR lags behind and predominantly focuses on English~\citep{marelli-etal-2014-sick, abdalla-etal-2023-makes}, mainly due to the lack of datasets. To close this gap, the SemEval-2024 Task 1: Semantic Textual Relatedness~\citep{ousidhoum-etal-2024-semeval} is proposed to encourage STR research on 14 African and Asian languages. The shared task consists of 3 subtasks: supervised (subtask A), unsupervised (subtask B), and cross-lingual (subtask C).\looseness-1

In this paper, we present our system AAdaM (\underline{A}ugmentation and \underline{Ada}ptation for \underline{M}ultilingual STR) developed for subtask A and C. Our system adopts a cross-encoder architecture which takes the concatenation of a pair of sentences as input and predicts the relatedness score through a regression head~\citep{devlin-etal-2019-bert}. As the provided task data for non-English languages is relatively limited, we perform data augmentation for these languages via machine translation. To better adapt a pre-trained model to the STR task, we apply task-adaptive pre-training~\citep{gururangan-etal-2020-dont} which has shown effectiveness on many tasks~\citep{xue-etal-2021-mt5, wang-etal-2023-nlnde}. For subtask A, we explore full fine-tuning and adapter-based tuning~\citep{pmlr-v97-houlsby19a} combined with previously mentioned techniques. Additionally, we use the adapter framework MAD-X~\citep{pfeiffer-etal-2020-mad} for cross-lingual transfer in subtask C. 

We select the best model based on the performance on development sets for the final submission, and our system achieves competitive results on both subtasks. In subtask A, our system ranks first out of 40 teams on average, and performs the best in Spanish. In subtask C, our system ranks first among 18 teams on average, and achieves the best performance in Indonesian and Punjabi.

\section{SemRel Dataset}
\label{sec:semrel}

\begin{figure}[]
    \centering
    \includegraphics[width=\columnwidth]{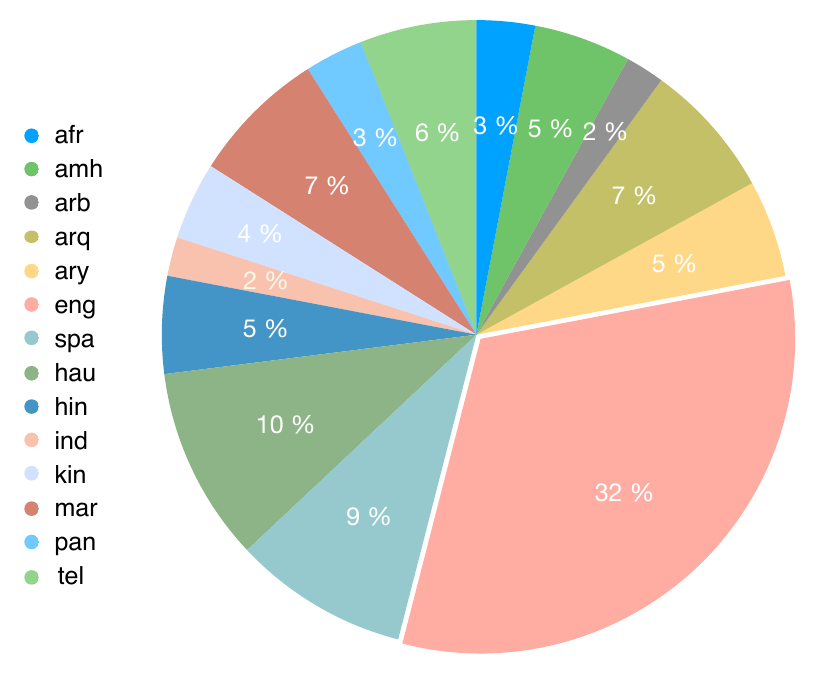}
    \caption{SemRel data distribution across languages.}
    \label{fig:data_distribution}
    \vspace{-5mm}
\end{figure}

To encourage STR research in the multilingual context, \citet{ousidhoum2024semrel2024} introduce SemRel, a new STR dataset annotated by native speakers, covering 14 languages from 5 distinct language families. These languages are mostly spoken in Africa and Asia, and many of them are under-represented in natural language processing resources. As shown in Figure~\ref{fig:data_distribution}, the data sizes vary widely from language to language constrained by the availability of resources. Notably, English data comprises 32\% of the whole dataset and surpasses other languages by a large margin.

\raggedbottom
\section{System Overview}
\label{sec:system}
Our system employs a \textit{cross-encoder} architecture, which takes the concatenation of a pair of sentences as input and predicts the relatedness score through a regression head. Compared to bi-encoders~\citep{reimers-gurevych-2019-sentence}, which extract individual sentence representations and then compare them using cosine similarity, cross-encoders generally perform better, at the cost of increased inference latency~\citep{Humeau2020Poly-encoders}. We select cross-encoder because of its superior performance (see Appendix~\ref{sec:appendix:arch}), and leave the exploration of an efficient alternative as future work.
\looseness-1

The core techniques underlying our system are (i) \textbf{data augmentation} using machine translation (\S\ref{sec:aug}), and (ii) \textbf{task-adaptive pre-training} on unlabeled task data (\S\ref{sec:tapt}). We explore two training paradigms for supervised learning combined with the aforementioned techniques, i.e., \textbf{fine-tuning} and \textbf{adapter-based tuning} (\S\ref{sec:tuning}), and the latter is also employed for cross-lingual transfer (\S\ref{sec:transfer}).

\raggedbottom
\subsection{Data Augmentation}
\label{sec:aug}

Data augmentation (DA) serves as a widely used strategy to mitigate data scarcity in low-resource languages~\citep{hedderich-etal-2021-survey,feng-etal-2021-survey}. Inspired by work on DA with machine translation~\citep{pmlr-v119-hu20b, amjad-etal-2020-data}, we create additional training data for non-English languages by translating from various English sources, as illustrated below.

\paragraph{SemRel translation.} As English data occupies a significant portion of the entire SemRel dataset, we perform augmentation by translating the English subset to other target languages. 
\paragraph{STS-B translation.} STS-B~\citep{cer-etal-2017-semeval}, a semantic similarity dataset, is highly relevant to STR, and we translate the STS-B training set in English to other target languages.  
\vspace{3mm}

\noindent It is worth noting that using translations as data augmentation yields a mixed data quality. For instance, the translation process may introduce artifacts that reduce data validity. Additionally, the concepts of ``similarity'' and ``relatedness'' are relevant but not equivalent, leading to a mismatch in their annotated scores. To leverage data in varied qualities, \citet{zhu-etal-2023-weaker} shows that a two-phase approach is beneficial, in which the model is trained on noisy data first and then trained on clean data. Our training procedure follows this two-phase scheme: (i) training the model on augmented data as a \textit{warmup}, and (ii) subsequently training the model on the original task data. 

\subsection{Task-Adaptive Pre-training}
\label{sec:tapt}
Pre-trained language models (PLMs) are trained on massive text corpora with self-supervision objectives for general purposes~\citep{devlin-etal-2019-bert, liu2019roberta}. To better adapt PLMs to downstream tasks, \citet{gururangan-etal-2020-dont} propose task-adaptive pre-training (TAPT), i.e., continued pre-training on task-specific unlabeled data, and show that it can effectively improve downstream task performance. We integrate this strategy into our system, wherein we conduct masked language modeling (MLM) on unlabeled task data for a given target language before initiating any supervised training.\looseness-1

\definecolor{Color}{rgb}{0.9, 0.9, 0.9}

\begin{table*}[t]
 \begin{center}
 \resizebox{\linewidth}{!}{
  \begin{tabular}{l|ll|ccccccccc}
    \toprule
    \textbf{Model Tuning}  &\textbf{TAPT} & \textbf{Warmup} &  \textbf{arq} & \textbf{amh} & \textbf{eng} & \textbf{hau} & \textbf{kin} & \textbf{mar} & \textbf{ary} & \textbf{spa} & \textbf{tel}  \\
    \midrule
    \midrule
    \multirow{6}{*}{\textsc{fine-tuning}}
    &\ding{55} & \ding{55}& 52.96 & 87.70 & 83.07 & 78.91 & 68.59 & 85.23 & 88.26 & \underline{73.83} & 84.90 \\
    &  \ding{55}& SemRel& 55.96 & 87.86 & / & 79.87 & 70.06 & 85.51 & \underline{\textbf{88.59}} & 72.93 & 85.38 \\
    &\ding{55}  & STS-B & 62.05 & 88.50 & \underline{\textbf{84.31}} & 79.86 & 69.78 & \underline{86.48} & 86.97 & 73.33 & 85.15 \\
    &\ding{51} & \ding{55} &65.70 & 88.03 & 82.79 & 79.41 & 67.03 & 84.88 & 88.50 & 70.47 & 83.84 \\
     & \ding{51}& SemRel& 66.74 & 85.58 & / & \underline{\textbf{80.73}} & \underline{71.29} & 85.74 & 87.01 & 73.37 & \underline{\textbf{85.77}} \\
    &\ding{51}& STS-B& \underline{\textbf{68.25}} & \underline{\textbf{88.72}} & 83.01 & 78.95 & 69.38 & 85.26 & 87.07 & 73.50 & 84.66 \\
    \midrule
    \multirow{6}{*}{\textsc{adapter tuning}}
    &\ding{55}  &\ding{55}  & 55.44 & 87.01 & 82.96 & 78.23 & 70.45 & 84.62 & 86.43 & 72.62 & 84.51 \\
    &\ding{55}& SemRel & 59.58 & \underline{87.66} & / & 79.15 & 70.56 & 86.54 & 86.88 & 74.90 & 84.88 \\
    &\ding{55}& STS-B& \underline{62.83} & 87.63 & \underline{82.97} & \underline{80.29} & 82.01 & 87.18 & 87.53 & 74.18 & 84.17 \\
    &\ding{51}& \ding{55} & 58.81 & 85.61 & 82.74 & 78.40 & 70.48 & 84.56 & 85.78 & 72.15 & 84.34 \\
    &\ding{51}& SemRel& 58.47 & 87.57 & / & 79.78 & 71.67 & \underline{\textbf{87.24}} & \underline{87.35} & \underline{\textbf{76.65}} & \underline{85.69} \\
    &\ding{51}& STS-B& 59.58 & 87.40 & 82.32 & 79.22 & \underline{\textbf{73.04}} & 87.12 & 87.22 & 73.22 & 83.70 \\
    \bottomrule
 \end{tabular}}
  \caption{Subtask A performance on development sets (Spearman's correlation $\times 100$). SemRel: warmup by training on SemRel translations; STS-B: warmup by training on STS-B translations. We \underline{underline} the best performance of fine-tuning and adapter-based tuning, and \textbf{bold} the best performance across all variants.}
  \label{tab:track_a_dev}
  \vspace{-5mm}
\end{center}
\end{table*}

\subsection{Fine-tuning vs. Adapter-based Tuning}
\label{sec:tuning}

Fine-tuning is the conventional approach to adapt general-purpose PLMs to downstream tasks. It updates all model parameters for each task, leading to inefficiency with the ever-increasing model scales and number of tasks. Recently, many works focus on introducing lightweight alternatives to improve parameter efficiency~\citep{lester-etal-2021-power, hu2022lora, he2022towards}. For example, adapter-based tuning~\citep{pmlr-v97-houlsby19a} only updates small modules known as adapters inserted between the layers of PLMs while keeping the remaining parameters frozen. In particular, it has shown impressive performance in cross-lingual transfer~\citep{pfeiffer-etal-2020-mad, ansell-etal-2021-mad-g, pfeiffer-etal-2022-lifting}.\looseness-1

We explore both fine-tuning and adapter-based tuning to compare their effectiveness on multilingual STR. For fine-tuning, we update all model parameters at each stage, namely the TAPT stage, the warmup stage and the final training stage using the original task data. For adapter-based tuning, we utilize the MAD-X framework~\citep{pfeiffer-etal-2020-mad} which consists of language-specific adapters and task-specific adapters. The language adapters are pre-trained with an MLM objective on unlabeled monolingual corpora. To this end, we collect open-source data from the Leipzig Corpus Collection~\citep{goldhahn-etal-2012-building} for pre-training.\footnote{Details are provided in Appendix~\ref{sec:appendix:data}.} The task adapters are trained on labeled task-specific data (augmented or original), while keeping the language adapters fixed. Note that when applying TAPT, only language adapters are updated. In subtask A, we apply fine-tuning and adapter-based tuning in combination with TAPT and warmup techniques, and select the best model based on the performance on development sets.

\subsection{Cross-lingual Transfer with Adapters}
\label{sec:transfer}
The high modularity of MAD-X enables efficient zero-shot cross-lingual transfer. During inference, we simply replace the source language adapter with the \textit{target language adapter} while retaining the \textit{source task adapter}. This task adapter has been trained on labeled data from the source language, without prior exposure to the target language.\footnote{Note that when transferring from any other language to English, we ensure that the source task adapter has not been trained on augmented data translated from English resources, thereby eliminating the effect of data leakage.} A crucial challenge for cross-lingual transfer lies in source language selection, as improper sources may lead to negative results~\citep{lange-etal-2021-share}. To determine the best source language, we explore the following metrics to rank sources: (1) linguistic distance~\citep{littell-etal-2017-uriel}, (2) token overlap~\citep{wu-dredze-2019-beto}, and (3) development set performance.\footnote{The existence of development sets is not realistic in the true zero-shot scenario, and we leave further discussion to the Limitations section.} Results in Appendix~\ref{sec:appendix:transfer} demonstrate that development set performance serves as the most reliable indicator of transfer performance. For subtask C, we select the optimal source from the adapters trained in subtask A based on their performance on development sets.

\definecolor{Color}{RGB}{230, 240, 249}

\begin{table*}[t]
 \begin{center}
 \resizebox{0.9\linewidth}{!}{
  \begin{tabular}{l|ccccccccc|c}
    \toprule
    \textbf{Model} & \textbf{arq} & \textbf{amh} & \textbf{eng} & \textbf{hau} & \textbf{kin} & \textbf{mar} & \textbf{ary} & \textbf{spa} & \textbf{tel} & \textbf{Avg.$\uparrow$} \\
    \midrule
    \midrule
    Overlap$^\diamondsuit$ & 40. & 63. & 67. & 31. & 33. & 62. & 63. & 67. & 70. & 55.11 \\
    LaBSE$^\diamondsuit$ & 60. & 85. & 83. & 69. & 72. & 88. & 77. & 70. & 82. & 76.22 \\
    \midrule
    PALI & 67.88 & \textbf{88.86} & \textbf{86.00} & \textbf{76.43} & 81.34 & \textbf{91.08} & \textbf{86.26} & 72.38 & 86.43 & \textbf{81.85} \\
    king001 & \textbf{68.23} & 88.78 & 84.30 & 74.72 & \textbf{81.69} & 89.68 & 85.97 & 72.12 & 85.34 & 81.20 \\
    NRK   & 67.36 & 86.42 & 83.29 & 67.20 & 75.69 & 87.93 & 82.70 & 68.99 & 83.42 & 78.11 \\
    saturn & 57.77 & 84.51 & - & 69.91 & 75.53 & 87.28 & 79.77 & - & \textbf{87.34} & - \\
    \rowcolor{Color}AAdaM (Ours) & 66.23 & 86.71 & 84.84 & 72.36 & 77.91 & 89.43 & 83.50 & \textbf{74.04} & 84.77 & 79.98 \\
    \bottomrule
  \end{tabular}}
  \caption{Subtask A performance on test sets (Spearman's correlation $\times 100$). $\diamondsuit$: baseline results from \citet{ousidhoum2024semrel2024}. We \textbf{bold} the best performance across submitted systems.}
  \label{tab:track_a_test}
  \vspace{-5mm}
\end{center}
\end{table*}

\begin{figure*}[]
    \centering
    \includegraphics[width=\linewidth]{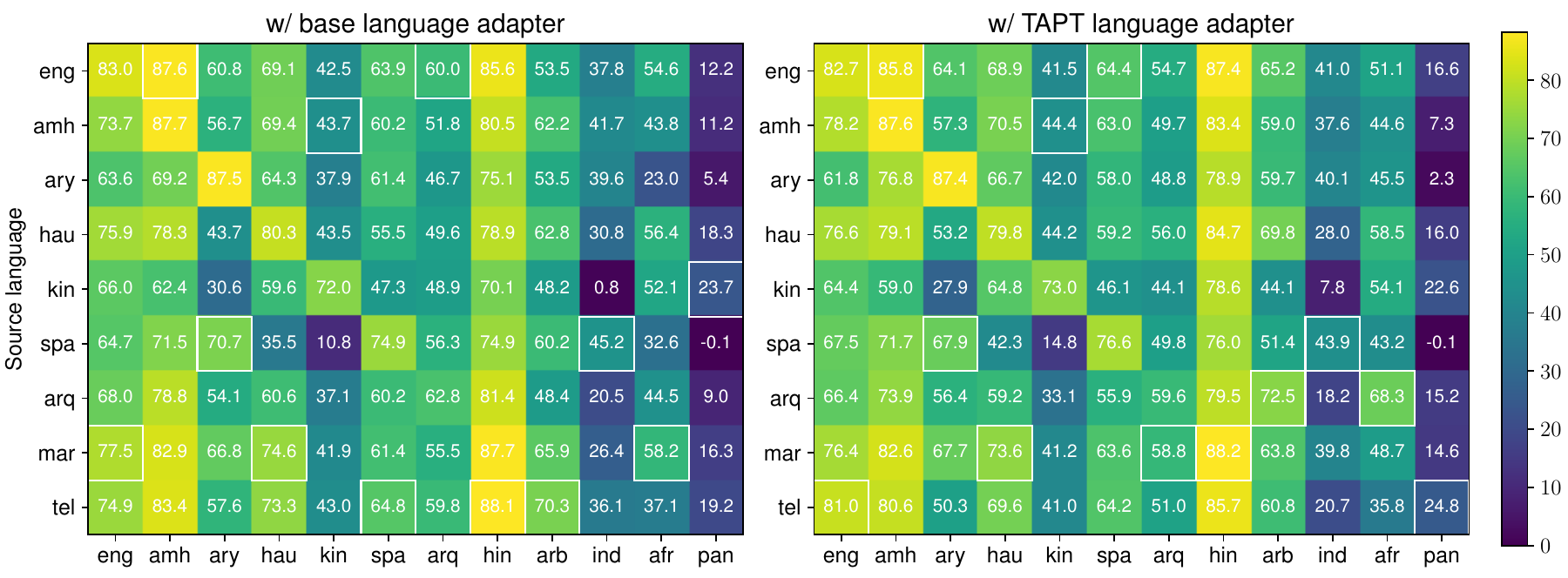}
    \vspace{-2em}
    \caption{Subtask C performance on development sets (Spearman's correlation $\times 100$) using different types of language adapters. Boxes highlight the optimal performances for each target language, and we select the best source for final submission.}
    \label{fig:transfer}
    \vspace{-5mm}
\end{figure*}

\section{Experimental Setup}
\paragraph{Model.} Our backbone model is AfroXLMR-large-61L~\citep{adelani-etal-2024-sib}, adapted from XLM-R~\citep{conneau-etal-2020-unsupervised} through multilingual adaptive fine-tuning~\citep{alabi-etal-2022-adapting}. We use NLLB (\texttt{nllb-200-distilled-600M})~\citep{nllbteam2022language} to translate from English resources to other languages as data augmentation.

\paragraph{Implementation.} All experiments are conducted on a single NVIDIA A100 GPU with a batch size of 16. For MLM, we set the learning rate to 5e-5 and train models for 10 epochs. For fine-tuning, we conduct a grid-search of learning rate from \{2e-5, 5e-5\} on SemRel development sets and train models for 6 epochs. For adapter-based tuning, we select the optimal learning rate from \{1e-4, 2e-4, 5e-5\} and train adapters for 15 epochs.

\definecolor{Color}{RGB}{230, 240, 249}

\begin{table*}[t]
 \begin{center}
 \resizebox{\linewidth}{!}{
  \begin{tabular}{l|cccccccccccc|c}
    \toprule
    \textbf{Model} & \textbf{afr} & \textbf{arq}  & \textbf{amh} & \textbf{eng} & \textbf{hau} & \textbf{hin} & \textbf{ind} & \textbf{kin} &   \textbf{arb} & \textbf{ary} & \textbf{pan} & \textbf{spa} & \textbf{Avg.$\uparrow$} \\
    \midrule
    \midrule
    Overlap$^\diamondsuit$ & 71. & 40. & 63. & 67. & 31. & 53. & 55. & 33. & 32. & 63. & -27. & 67. & 45.67 \\
    LaBSE$^\diamondsuit$ & 79. & 46. & 84. & 80. & 62. & 76. & 47. & 57. & 61. & 40. & -5. & 62. &  57.42\\
    \midrule
    king001 & 81.00 & \textbf{61.44} & \textbf{87.83} & -   & \textbf{73.35} & \textbf{84.39} & 37.58 & 62.99 & 65.68 & \textbf{81.96} & -   & \textbf{70.76} & - \\
    UAlberta & 80.57 & 44.13 & 81.60 & -   & 67.85 & 82.78 & 44.90 & 63.58 & \textbf{67.15} & 60.22 & -1.74 & 57.16 & - \\
    ustcctsu & 74.87 & 41.44 & 70.90 & 78.40 & 47.63 & 65.80 & 46.02 & 45.41 & 46.87 & 61.32 & -24.79 & 68.51 & 51.87 \\
    umbclu & \textbf{82.23} & 12.63 & 4.30  & 78.75 & 45.69 & 15.52 & 51.53 & 48.36 & 3.54  & -3.75 & -7.75 & 60.89 & 32.66 \\
    \rowcolor{Color}AAdaM (ours)  & 81.39 & 55.07 & 86.29 & \textbf{79.37} & 72.88 & 83.86 & \textbf{52.80} & \textbf{64.99} & 65.32 & 60.03 & \textbf{15.53} & 62.05 & \textbf{64.97} \\
    \bottomrule
 \end{tabular}}
  \caption{Subtask C performance on test sets (Spearman's correlation $\times 100$). $\diamondsuit$: baseline results from \citet{ousidhoum2024semrel2024}. We \textbf{bold} the best performance across submitted systems.}
  \label{tab:track_c_test}
  \vspace{-5mm}
\end{center}
\end{table*}

\section{Results and Analysis}
\label{sec:results}

\subsection{Subtask A: Supervised Learning}
In Table~\ref{tab:track_a_dev}, we compare the performance on development sets using fine-tuning and adapter-based tuning along with various techniques. Fine-tuning achieves the best performance in most languages (6 out of 9), which is unsurprising as it optimizes the entire parameter space. Notably, adapter-based tuning demonstrates comparable performance to fine-tuning in Hausa (hau) and Telugu (tel), while even surpassing it in Kinyarwanda (kin), Marathi (mar) and Spanish (spa). Looking at the effectiveness of TAPT and warmup, we observe that they provide benefits in most cases compared to using no techniques at all. Nonetheless, the improvements are sometimes marginal, particularly in languages such as Amharic (amh), English (eng), and Moroccan Arabic (ary), where the baseline performances are already relatively strong compared to other languages.

In our final submission, we selected the best model for each language based on the performance of development sets. As shown in Table~\ref{tab:track_a_test}, our approach largely improves the baseline results~\citep{ousidhoum2024semrel2024}, especially for Algerian Arabic (arq), Kinyarwanda (kin), and Moroccan Arabic (ary). In comparison to several top-performing submitted systems, we achieve the best performance in Spanish (spa). There were a total of 40 final submissions in subtask A, and our system ranks first on average in the official leaderboard.\footnote{PALI and king001 also achieved competitive performance; however, they are not ranked in the official leaderboard due to missing system descriptions.}

\subsection{Subtask C: Cross-lingual Transfer}
In subtask C, we replace source language adapters from subtask A with target language adapters. We analyze two groups of language adapters: base language adapters trained only on Leipzig corpora and TAPT language adapters further trained on unlabeled task data. The cross-lingual transfer results on development sets are shown in Figure~\ref{fig:transfer}. We observe a discrepancy in the optimal source languages selected with two types of adapters, indicating a behavior shift after applying TAPT. 
Furthermore, the performance for target languages shows high sensitivity to the choice of source language. For example, using Spanish (spa) as the source language for Indonesian (ind) performs significantly better than using Kinyarwanda (kin), showcasing the importance of careful source language selection. When examining each target language, we find that in the case of Amharic (amh), the cross-lingual transfer performance is comparable to its supervised learning performance. However, it remains a challenge for a few languages, such as Indonesian (ind) and Punjabi (pan). 

The results for test sets are shown in Table~\ref{tab:track_c_test}. Compared to LaBSE~\citep{feng-etal-2022-language}, a multilingual sentence embedding model, our cross-lingual transfer approach achieves better performance on most languages, especially for Algerian Arabic (arq), Hausa (hau), Moroccan Arabic (ary), and Punjabi (pan). However, our system is surpassed by the simple word overlap baseline in Indonesian (ind), Moroccan Arabic (ary) and Spanish (spa). This highlights the need for nuanced investigation of data distributions across various languages. Subtask C received 18 submissions in total, and we perform the best in the official leaderboard. In particular, we achieve the best performance in Indonesian (ind) and Punjabi (pan), which seem harder for other teams. For Punjabi (pan), where most teams get negative correlation scores, our method maintains its effectiveness.\looseness-1

\subsection{Analysis}
\begin{figure}[]
    \centering
    \includegraphics[width=\columnwidth]{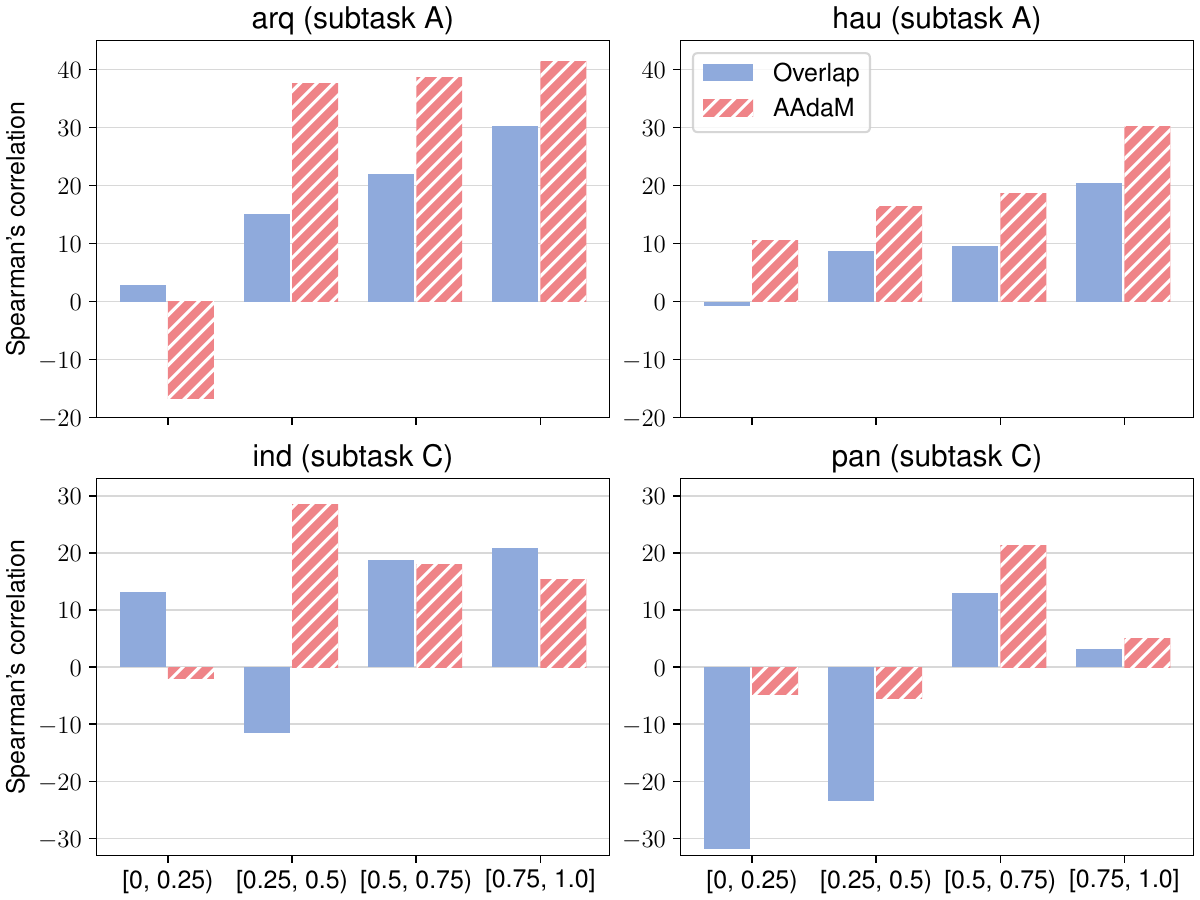}
    \vspace{-1.5em}
    \caption{Performance on test sets (Spearman's correlation $\times 100$) in different relatedness levels.}
    \label{fig:analysis}
    \vspace{-2mm}
\end{figure}

We partition ground-truth relatedness scores, ranging from 0 to 1, to different levels for fine-grained analysis. Figure~\ref{fig:analysis} shows the detailed model performance for several under-performing languages. Although our evaluation scores on the entire test sets are all positive, some subsets exhibit negative correlations, particularly those with lower relatedness scores. Moreover, AAdaM largely lags behind the simple word overlap baseline for Algerian Arabic (arq) and Indonesian (ind) within the 0 to 0.25 range. These observations highlight the complexity of capturing nuanced relationships within specific categories, possibly affected by the data annotation procedure and unbalanced learning. 

\section{Conclusion}
In this paper, we introduce our multilingual STR system, AAdaM, developed for the SemEval-2024 Task 1, which achieves competitive results in both subtask A and subtask C. We see noticeable improvements by using data augmentation and task-adaptive pre-training, and demonstrate that adapter-based tuning is an effective approach for supervised learning and cross-lingual transfer. Despite these strengths, our fine-grained analysis reveals that capturing nuanced semantic relationships remains a challenge, highlighting the need for further granular investigation and modeling improvements.

\clearpage
\section*{Limitations}
Although our approach has demonstrated impressive performance, relying on development sets for source language selection undermines its practical value in the true zero-shot setting. While linguistic (dis)similarity~\citep{littell-etal-2017-uriel} is a commonly used estimator for cross-lingual transfer performance, it alone does not explain many transfer results~\citep{lauscher-etal-2020-zero}. \citet{philippy-etal-2023-towards} survey different factors that impact cross-lingual transfer performance, finding contradictory conclusions from previous studies. In future work, we plan to scrutinize the interplay among various factors, and select the optimal source language without relying on post-hoc evaluation.

\section*{Acknowledgements}
We thank Vagrant Gautam and Badr M. Abdullah for their proofreading and anonymous reviewers for their feedback. Miaoran Zhang received funding from the DFG (German Research Foundation) under project 232722074, SFB 1102. Jesujoba O. Alabi was supported by the BMBF’s (German Federal Ministry of Education and Research) SLIK project under the grant 01IS22015C.

\bibliography{anthology,custom}
\bibliographystyle{acl_natbib}

\appendix

\section{Model and Architecture Selection}
\label{sec:appendix:arch}

In our preliminary study, we examine the capacity of different pre-trained models with or without any training. To assess their out-of-the-box effectiveness, we extract contextual embeddings for pairs of sentences from various multilingual models, and use the cosine similarity to predict the semantic relatedness score. The multilingual models include: 

\begin{itemize}
    \item \textbf{sentence transformers}: mpnet-base-v2\footnote{\url{https://huggingface.co/sentence-transformers/paraphrase-multilingual-mpnet-base-v2}} and LaBSE~\citep{feng-etal-2022-language}
    \item \textbf{general-purpose models}: XLMR-large~\citep{conneau-etal-2020-unsupervised}, AfroXLMR-large~\citep{alabi-etal-2022-adapting}, AfriBERTa-large~\citep{ogueji-etal-2021-small},  AfroXLMR-large-61L and AfroXLMR-large-75L~\citep{adelani-etal-2024-sib}
\end{itemize}

Additionally, we add two simple baselines for comparison: word overlap\footnote{\url{https://github.com/semantic-textual-relatedness/Semantic_Relatedness_SemEval2024/blob/main/STR_Baseline.ipynb}} and fastText~\citep{mikolov-etal-2018-advances}. For both fastText vectors and contextual embeddings, we employ mean pooling to get sentence embeddings. 

In Table~\ref{tab:model_selection}, we can see that sentence transformers achieve superior performance in most languages when no training is conducted. This observation is not unsurprising, as they have been trained for sentence embeddings that can better capture the semantic relationships. However, this trend shifts upon fine-tuning the models on task data with either bi-encoder or cross-encoder architecture. Notably, with the cross-encoder architecture, AfroXLMR-large-61L achieves comparable performance to LaBSE. To satisfy the requirement in subtask C, for which the pre-trained model should not be trained on any relatedness or similarity datasets, we adopt AfroXLMR-large-61L as our backbone model with the cross-encoder architecture for all our experiments.

\begin{table*}[t]
\begin{center}
\resizebox{\textwidth}{!}{
\begin{tabular}{l|llllllll|l}
\toprule
\textbf{Model} & \textbf{eng} & \textbf{amh}   & \textbf{arq}   & \textbf{ary}   & \textbf{spa}   & \textbf{hau}   & \textbf{mar}   & \textbf{tel}   & \textbf{Avg.$\uparrow$}  \\ 
\midrule
\midrule
\multicolumn{10}{l}{\textbf{\textit{Baselines w/o training:}}} \\
Overlap & 56.57   & 63.28 & 44.00 & \textbf{53.76} & 58.67 & 38.86 & 57.52 & 60.61 & 54.16 \\
FastText  & 55.69 & 60.64 & \textbf{44.27} & 22.12 & 57.47 & 9.19  & 59.23 & 69.39 & 47.25 \\
\cdashlinelr{1-10}
mpnet-base-v2  & \textbf{81.94} & 69.94 & 26.35 & 34.40  & 56.58 & 30.86 & 72.43 & 56.33 & 53.60 \\
LaBSE & 72.14  & \textbf{76.49} & 40.80  & 38.58 & \textbf{63.11} & \textbf{41.51} & \textbf{73.83} & \textbf{75.99} & \textbf{60.31} \\
XLMR-large & 39.53 & 42.07 & 27.91 & 4.15  & 47.59 & 7.34  & 40.51 & 56.36 & 33.18 \\
AfroXLMR-large   & 16.55  & 39.82 & 20.30  & -0.46 & 30.42 & 8.13  & 35.94 & 30.74 & 22.68 \\
AfriBERTa-large   & 53.12  & 69.23 & 16.04 & 13.36 & 56.68 & 35.14 & 20.84 & 9.73  & 34.27 \\
AfroXLMR-large-61L  & 44.10 & 52.96 & 32.15 & 0.35  & 51.07 & 17.62 & 37.66 & 47.17 & 35.39 \\
AfroXLMR-large-75L  & 22.61  & 37.93 & 29.38 & -2.39 & 43.58 & 13.86 & 32.13 & 40.42 & 27.19 \\
\midrule
\multicolumn{10}{l}{\textbf{\textit{Bi-encoders w/ supervised training:}}} \\
mpnet-base-v2  & 85.07  & 80.43 & 56.73 & 75.51 & 65.29 & 58.62 & 81.53 & 74.49 & 72.21 \\
LaBSE  & 84.45 & 82.59 & 59.49 & 78.29 & 69.02 & 68.94 & \textbf{83.97} & 76.35 & 75.39 \\
AfroXLMR-large-61L & 82.81 & 74.61 & 40.02 & 66.58 & 66.65 & 66.51 & 38.51 & 65.73 & 62.68 \\ 
\cdashlinelr{1-10}
\multicolumn{10}{l}{\textbf{\textit{Cross-encoders w/ supervised training:}}} \\
mpnet-base-v2  & 80.26  & 75.04 & 60.25 & 80.31 & 64.92 & 53.66 & 65.36 & 68.54 & 68.54 \\
LaBSE  & 86.13  & 84.75 & \textbf{60.75} & \textbf{82.55} & 67.23 & 69.31 & 81.10  & 77.25 & \textbf{76.13} \\
AfroXLMR-large-61L & \textbf{86.65} & \textbf{84.88} & 46.61 & 81.56 & \textbf{69.08} & \textbf{74.65} & 75.55 & \textbf{80.94} & 74.99 \\
\bottomrule
\end{tabular}}
\caption{Performance of 10-fold cross-validation on training sets (Spearman's correlation $\times 100$). For each language, we \textbf{bold} the best performance achieved in \textit{w/o training} and \textit{w/ supervised training} settings.}
\label{tab:model_selection}
\end{center}
\end{table*}

\begin{table*}[!hbt]
 \begin{center}
 \resizebox{\linewidth}{!}{
  \begin{tabular}{lllr}
    \toprule
    \textbf{Language} & \textbf{Family / Subfamily} & \textbf{Domain} & \textbf{Corpus Size} \\
    \midrule
    English (eng) & Indo-Europoean / Germanic & News, Wikipedia & 1.2M \\
    Afrikaans (afr) & Indo-Europoean / Germanic & News, Wikipedia & 68k \\
    Amharic (amh) & Afro-Asiatic / Semitic &Community, Wikipedia & 250k \\
    Modern Standard Arabic (arb) & Afro-Asiatic / Semitic &News, Wikipedia & 110k \\
    Algerian Arabic (arq) &Afro-Asiatic / Semitic & News & 244k \\
    Moroccan Arabic (ary) & Afro-Asiatic / Semitic &News & 564k\\
    Spanish (spa) & Indo-Europoean / Italic &News, Wikipedia & 444k \\
    Hausa (hau) & Afro-Asiatic / Chadic & Community, Wikipedia & 564k \\
    Hindi (hin) & Indo-European / Indo-Iranian & News, Wikipedia & 472k\\
    Indonesian (ind) & Austronesian / Malayic &News, Wikipedia & 92k \\
    Kinyarwanda (kin) & Niger-Congo / Atlantic–Congo &Community & 320k \\
    Punjabi (pan) & Indo-European / Indo-Iranian &Wikipedia & 412k \\
    Marathi (mar) & Indo-European / Indo-Iranian &News, Wikipedia & 856k \\
    Telugu (tel) & Dravidian / South-Central &News, Wikipedia & 756k\\
    \bottomrule
 \end{tabular}}
  \caption{Data statistics for pre-training corpora collected from the Leipzig Corpus Collection.}
  \label{tab:leipzig}
\end{center}
\end{table*}

\section{Pre-training Data Collection}
\label{sec:appendix:data}
To pre-train language adapters, we collect open-source corpora from the Leipzig Corpus Collection and use the recent data derived from news and Wikipedia domains. Data statistics are shown Table~\ref{tab:leipzig}. As the SemRel data spans diverse domains, there is a potential risk of domain mismatch between the pre-training data and task data, which needs further investigation.

\section{Source Language Selection}
\label{sec:appendix:transfer}
To determine the best source language for cross-lingual transfer, we explore three metrics to estimate the transfer performance:

\paragraph{Linguistic distance.} We use the average of six distances obtained from the URIEL database~\citep{littell-etal-2017-uriel} to measure the similarity between a pair of languages. These distances include syntactic, phonological, inventory, geographic, genetic, and featural distances. A lower distance indicates that the two languages are more similar, potentially facilitating more effective transfer.

\paragraph{Token overlap.} We follow~\citep{wu-dredze-2019-beto} to measure how many tokens are shared in the source training set and the target test set. A higher token overlap indicates that more tokens were encountered during training in the source language, potentially transferring more supervision from the source to the target.  

\paragraph{Development set performance.} As small development sets are available in the shared task, we use their performance as an indicator of the transfer performance on test sets, assuming that they share a similar data distribution.\footnote{When training is allowed, it might be more advantageous to use small development sets for training directly rather than source selection, which needs to be further explored.}

\begin{figure*}[!htb]
     \centering
     \begin{subfigure}[b]{\linewidth}
         \centering
         \includegraphics[width=\linewidth]{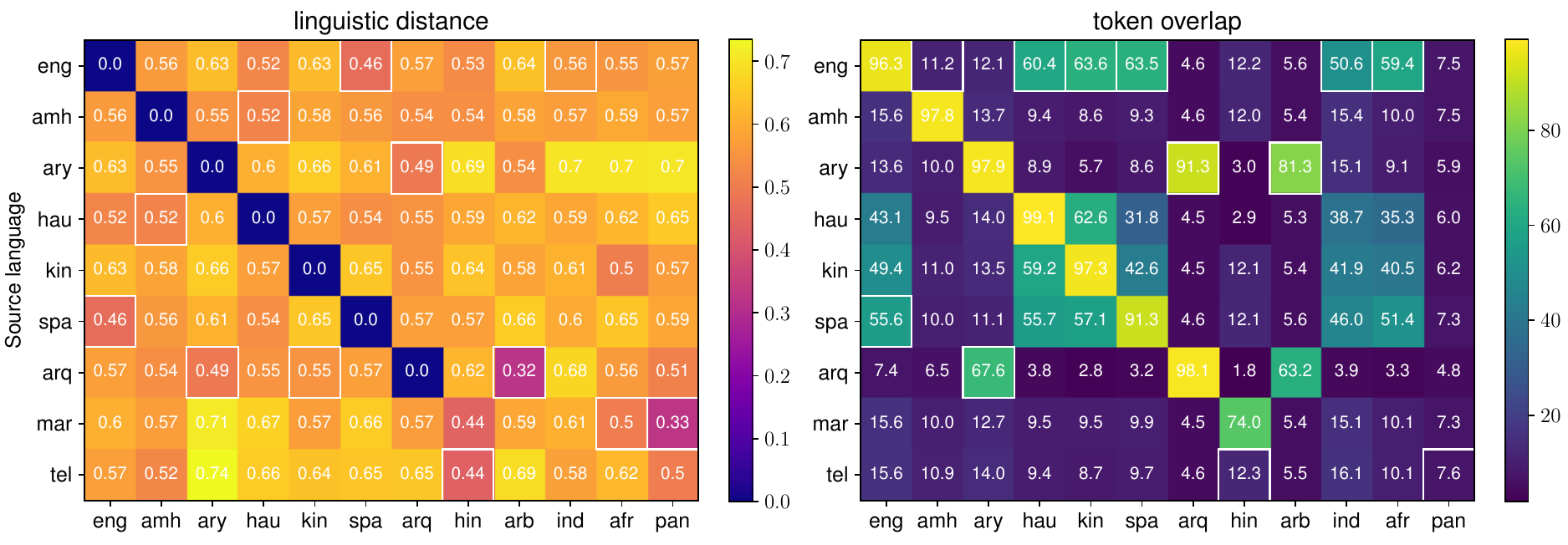}
         \caption{Left: Linguistic distances between source and target languages. The smallest distance for each target language is highlighted with a box.
          Right: Token overlaps between source and target languages. The highest overlap for each target language is highlighted with a box. The corresponding source languages are predicted as the best sources for cross-lingual transfer.}
         \label{fig:transfer_pred}
     \end{subfigure}
     \hfill
     \begin{subfigure}[b]{\linewidth}
         \centering
         \includegraphics[width=\linewidth]{figures/transfer_dev.pdf}
         \caption{Performance on development sets (Spearman’s correlation ×100) using different types of
            language adapters. Boxes are used to highlight the optimal performances for each target language, and the corresponding source languages are predicted as the best sources for cross-lingual transfer.}
         \label{fig:transfer_dev}
     \end{subfigure}
     \hfill
     \begin{subfigure}[b]{\linewidth}
         \centering
         \includegraphics[width=\linewidth]{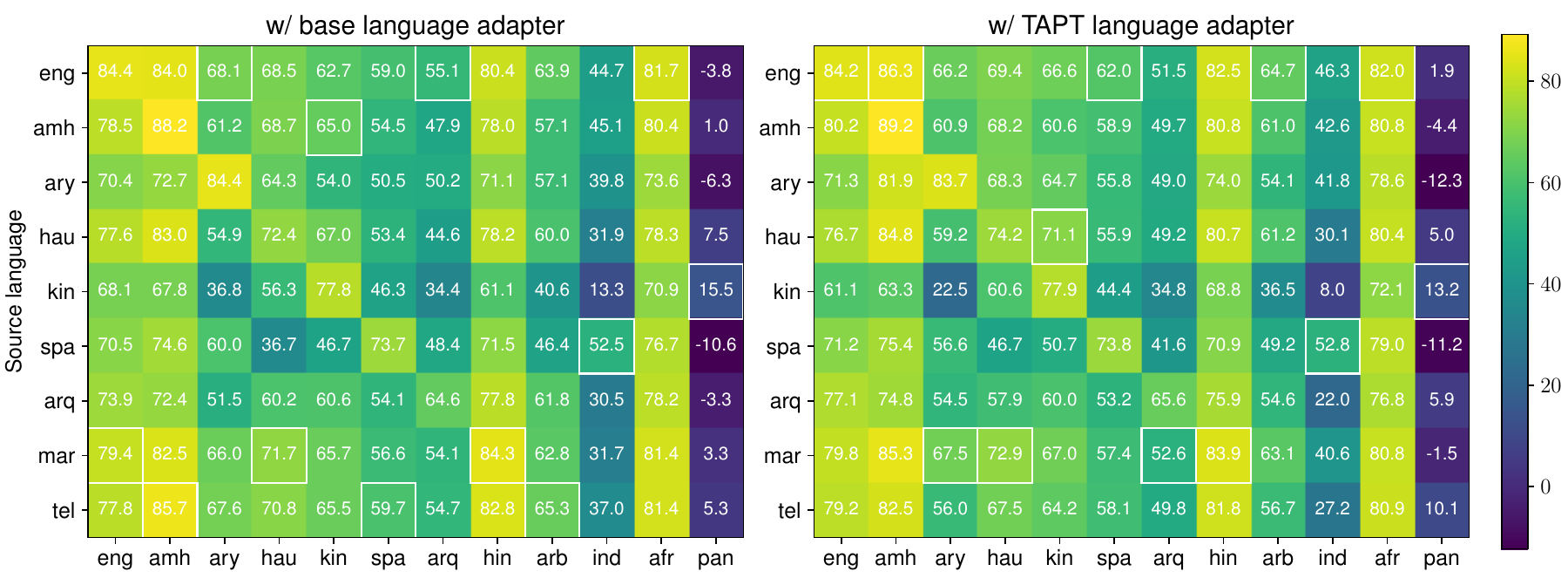}
         \caption{Performance on test sets (Spearman’s correlation ×100) using different types of
            language adapters. Boxes are used to highlight the optimal performances for each target language, and the corresponding source languages are the ground-truth best sources for cross-lingual transfer.}
         \label{fig:transfer_test}
     \end{subfigure}
     \caption{Comparison of different source language selection methods.}
     \label{fig:transfer_pred_dev_test}
\end{figure*}

\vspace{3mm}
In Figure~\ref{fig:transfer_pred_dev_test}, we show the metric values across different source languages, along with the best source languages identified by distinct metrics. After post-hoc evaluation following the release of test sets, we find that the performance of the development set indeed serves as the most reliable indicator, as the optimal source languages it selected closely align with the ground truth selections. 

\end{document}